\documentclass[conference]{IEEEtran}
\IEEEoverridecommandlockouts
\usepackage{cite}
\usepackage{amsmath,amssymb,amsfonts}
\usepackage{algorithmic}
\usepackage{graphicx}
\usepackage{textcomp}
\usepackage{xcolor}
\usepackage{xurl}
\usepackage{hyperref}

\def\BibTeX{{\rm B\kern-.05em{\sc i\kern-.025em b}\kern-.08em
    T\kern-.1667em\lower.7ex\hbox{E}\kern-.125emX}}

\begin{document}

\title{Exploring the Impact of Outlier Variability on Anomaly Detection Evaluation Metrics
}
% \author{\IEEEauthorblockN{Anonymous Authors}}
\author{\IEEEauthorblockN{1\textsuperscript{st} Minjae Ok}
\IEEEauthorblockA{\textit{Department of Statistics} \\
\textit{TU Dortmund}\\
Dortmund, Germany \\
minjae.ok@tu-dortmund.de}
\and
\IEEEauthorblockN{2\textsuperscript{nd} Simon Klüttermann}
\IEEEauthorblockA{\textit{Department of Computer Science} \\
\textit{TU Dortmund}\\
Dortmund, Germany \\
simon.kluettermann@cs.tu-dortmund.de}
\and
\IEEEauthorblockN{3\textsuperscript{rd} Emmanuel Müller}
\IEEEauthorblockA{\textit{Department of Computer Science} \\
\textit{TU Dortmund}\\
Dortmund, Germany \\
emmanuel.mueller@cs.tu-dortmund.de}
}

\maketitle

\begin{abstract}
Anomaly detection is a dynamic field, in which the evaluation of models plays a critical role in understanding their effectiveness. The selection and interpretation of the evaluation metrics are pivotal, particularly in scenarios with varying amounts of anomalies. This study focuses on examining the behaviors of three widely used anomaly detection metrics under different conditions: F1 score, Receiver Operating Characteristic Area Under Curve (ROC AUC), and Precision-Recall Curve Area Under Curve (AUCPR). Our study critically analyzes the extent to which these metrics provide reliable and distinct insights into model performance, especially considering varying levels of outlier fractions and contamination thresholds in datasets. 

Through a comprehensive experimental setup involving widely recognized algorithms for anomaly detection, we present findings that challenge the conventional understanding of these metrics and reveal nuanced behaviors under varying conditions. We demonstrated that while the F1 score and AUCPR are sensitive to outlier fractions, the ROC AUC maintains consistency and is unaffected by such variability. Additionally, under conditions of a fixed outlier fraction in the test set, we observe an alignment between ROC AUC and AUCPR, indicating that the choice between these two metrics may be less critical in such scenarios.

The results of our study contribute to a more refined understanding of metric selection and interpretation in anomaly detection, offering valuable insights for both researchers and practitioners in the field.
\end{abstract}


\section{Introduction}
Anomaly detection, characterized by its dynamic and evolving nature, continually presents unique challenges for model evaluation. This process involves identifying rare or unusual events within datasets, which can provide critical insights or indicate potential issues within systems. The selection of appropriate evaluation metrics at the heart of these challenges~\cite{Sorbo2023Navigating} is critical for an accurate and meaningful understanding of a model's performance~\cite{goldstein2016comparative,lavin2015evaluating,ozturk2021adnet,khelifati2021vadetis}. Complicating this; different papers suggest using different metrics. While the Receiver Operating Characteristic Area Under Curve (ROC AUC)~\cite{ROC_AD} is the most commonly used metric, other studies\cite{surveyzhao} argue that the Precision-Recall Curve Area Under Curve (AUCPR)~\cite{AUCPR_AD} should be used and mostly in applied papers the F1 score~\cite{F1_AD} is commonly found. Our study focused on comparing these metrics and 
%this fundamental aspect and aimed to 
delves into the behaviors of them under diverse testing scenarios.% The primary focus lies on three commonly used metrics in anomaly detection: the F1 score~\cite{F1_AD}, Receiver Operating Characteristic Area Under Curve (ROC AUC)~\cite{ROC_AD}, and Precision-Recall Curve Area Under Curve (AUCPR)~\cite{AUCPR_AD}.

While most of our findings can be applied both to 
%Although anomaly detection can be approached using both
(semi-)supervised and unsupervised anomaly detection, we conduct our research in an unsupervised setting. This focus is justified, since most anomaly detection applications are unsupervised. This is because labeling possibly very rare anomalies can be very complex.%Instead, unsupervised anomaly detection, not depending on pre-labeled datasets, can provides a distinct landscape for exploring and understanding these metrics.

These metrics have been widely adopted because of their perceived distinctiveness in evaluating the model performance. However, their effectiveness and reliability, particularly under varying outlier fractions and contamination levels, have not been examined thoroughly. Our research is driven by the challenge of selecting the most appropriate metric from a set of available options, where each metric has its own unique strengths and weaknesses. In practical settings, where using multiple metrics simultaneously is often not feasible or confusing, determining the most suitable single metric becomes crucial. This study aims to provide a more nuanced understanding of these metrics, particularly in their response to different levels of anomaly presence in datasets. By analyzing the behavior of the F1 score, ROC AUC, and AUCPR across various scenarios, we seek to illuminate the key differences between these metrics. This detailed exploration will guide practitioners in making more informed decisions about metric selection. Ultimately, this work contributes to a clearer understanding of the utility and limitations of these metrics in the field of anomaly detection.

To complement our analysis of real-world datasets, this study employs simulated environments, which provide controlled means to model and adjust the mean separation between Gaussian-distributed classes. This approach allows us to extend our analysis by systematically exploring the sensitivity of evaluation metrics, specifically ROC AUC and AUCPR, to changes in the degree of separation between normal and anomalous instances. By integrating empirical analysis with simulations, we aim to comprehensively validate the metrics' robustness and effectiveness in anomaly detection across a spectrum of real-world and idealized scenarios.

%section explanation
In \textit{Section II: Related Work}, we explore the extensive research in anomaly detection algorithms and the ongoing debate on the effectiveness of various evaluation metrics. This section sets the stage for understanding the current landscape and the gaps our study aims to address.

\textit{Section III: Experimental Setup} describes our multifaceted experimental approach, detailing the algorithms used and the rationale behind the selection of specific evaluation metrics and contamination thresholds. This section elucidates the methodological framework of our study.

In \textit{Section IV: Methodology}, we delve into the dataset preparation, algorithm implementation, and the specific procedures for training and testing splits. This section provides a comprehensive overview of the practical aspects of our research.

\textit{Section V: Results and Discussion} presents the findings from our experiments, offering insights into the correlation patterns among the F1 score, ROC AUC, and AUCPR under different contamination levels and outlier fractions. This section is crucial for interpreting the implications of our research findings.

Finally, \textit{Section VI: Conclusion} summarizes the key findings and their significance in the broader context of anomaly detection, highlighting the contributions of our study to the field.

Our code is available at \url{https://anonymous.4open.science/r/AD_EM-F5D3/README.md}.

\section{Related Work}
Anomaly detection has undergone extensive research, leading to the development of various algorithms, each with unique approaches to identifying outliers. While our study does not primarily focus on these algorithms, it employs several, such as K-Nearest Neighbors (KNN)~\cite{knn}, Local Outlier Factor (LOF)~\cite{lof}, One-Class Support Vector Machine (OCSVM)~\cite{ocsvm}, and Isolation Forest (IForest)~\cite{isoforest}~\cite{goldstein2016comparative}. These algorithms serve as tools to generate diverse testing scenarios, crucial for our study's robust evaluation of metrics. The choice of these specific algorithms reflects their varied approaches to anomaly detection, ensuring a comprehensive testing ground for our analysis of evaluation metrics.

Simultaneously, evaluating these models remains a subject of ongoing debate, particularly regarding the choice and effectiveness of evaluation metrics~\cite{ozturk2021adnet}~\cite{Moghaddam.2019}. Central to this debate are the F1 score, ROC AUC, and AUCPR~\cite{inproceedings}. The selection of these metrics often depends on the specific aims and challenges of each study. The F1 score, valued for its balance between precision and recall, is commonly favored when both false positives and false negatives are equally important~\cite{f1score}. ROC AUC is known for its ability to assess a model's discriminative power, remaining unaffected by class imbalance and thereby offering robustness in diverse detection scenarios~\cite{rocauc}. AUCPR, on the other hand, is noted for its effectiveness in evaluating model performance in relation to the minority class, a significant consideration in highly imbalanced datasets~\cite{Keilwagen.2014}.

The ongoing debate in the field centers on the suitability and effectiveness of these metrics across different anomaly detection scenarios. Traditional views suggest that each metric provides unique insights tailored to specific conditions. However, our study challenges this perceived distinctiveness. We observe that the differences between metrics like the ROC AUC and AUCPR might be minimal under certain conditions, such as a consistent outlier fraction. This finding suggests that these metrics might align more closely in their evaluation outcomes than previously recognized, especially in scenarios with a stable outlier fraction. This evolving understanding necessitates a reexamination of metric selection, pushing the boundaries of current methodologies and advocating for a more nuanced approach in evaluating anomaly detection models.

% Our study builds upon these insights, aiming to investigate the behavior of these metrics under varying conditions. By exploring the impact of different contamination levels and outlier fractions, we seek to understand the extent to which these metrics can reliably represent model performance in anomaly detection.

% In summary, this paper situates our research within the broader context of anomaly detection, acknowledging the field's diversity of approaches and the evolving understanding of evaluation metrics. It underscores the need for continuous examination and adaptation of these metrics, to effectively meet the challenges posed by varying data characteristics and anomaly detection objectives.

\section{Experimental setup}
Building upon the foundational understanding of evaluation metrics in anomaly detection, as discussed in the related work section, our study employs a multifaceted experimental approach. The study utilized four widely-used anomaly detection algorithms: K-Nearest Neighbors (KNN), Local Outlier Factor (LOF), One-Class Support Vector Machine (OCSVM), and Isolation Forest (IForest). These algorithms were specifically selected for their varied methodologies in identifying outliers, encompassing distance-based (KNN), density-based (LOF), partitioning-based (IForest), or boundary-based (OCSVM). %It is important to note that our approach falls under the domain of unsupervised anomaly detection. Unlike supervised anomaly detection, our methods do not rely on pre-labeled datasets for training.

In line with the ongoing debate around the effectiveness of different evaluation metrics, our study focuses on three key measures: the F1 score, Receiver Operating Characteristic Area Under Curve (ROC AUC), and Precision-Recall Curve Area Under Curve (AUCPR). These metrics were chosen due to their prominence in the field and the varying insights they offer into model performance.

A key aspect of our experimental design, particularly relevant to the unsupervised nature of our anomaly detection approach, is the use of different contamination thresholds. It is crucial to note that the training set comprises solely normal instances in unsupervised anomaly detection, with no/neglectable anomalies included. This characteristic is typical of unsupervised methods, where anomaly labels are unavailable or identified during the training phase. Still we expect our results to generalize to other applications with highly imbalanced data.

As commonly done in anomaly detection tasks\cite{pyod}, we define contamination not as an indication of actual outliers present in the training data but as a means to set a uniform threshold across datasets. This threshold, which we refer to as the contamination threshold, represents the expected proportion of outliers in the data. For this, we accept certain levels of false positives. Specifically, we set the contamination thresholds at three distinct levels: $1\%$, $5\%$, and $10\%$. This approach allows us to uniformly apply a threshold across datasets while acknowledging and controlling for the expected rate of false positives.

Such a configuration is vital for assessing how the F1 score, a metric that balances precision and recall, reacts under different levels of assumed outlier presence or different chosen thresholds. By varying these contamination levels, our objective is to thoroughly understand the F1 score's responsiveness and its ability to reflect model performance in diverse scenarios accurately. Furthermore, it is important to note that, unlike the F1 score, ROC AUC and AUCPR do not require a predefined threshold. These metrics evaluate the model's performance across various thresholds, offering a more comprehensive assessment of performance. This can also be interpreted as them averaging over all possible thresholds.

Additionally, the outlier fraction in the test set was a central element of our experimental design, implemented in two distinct configurations: a fixed fraction and a random fraction. 

For the fixed outlier fraction configuration, the test set was composed of all anomalies in the dataset and an equal number of normal observations, while the remaining normal observations formed the training set. This setup ensures a controlled environment where the fraction of anomalies, or the outlier fraction, is consistently maintained at $50\%$. This fixed fraction setup is crucial for creating a predictable and stable testing environment, facilitating the evaluation of anomaly detection methods under controlled conditions.

Conversely, in the random fraction configuration, $30\%$ of the normal samples were randomly selected to form the test set, irrespective of how many anomalies were given. The remaining $70\%$ of the dataset formed the training set. This configuration reflects a more realistic scenario where the proportion of outliers in the test set is unknown and variable.

It is important to note that in the context of unsupervised anomaly detection, the presence or absence of anomalies in the training set is not a primary concern. The algorithms used in our study are designed to identify outliers without prior knowledge of their existence, making them effective even in scenarios where anomalies are included in the training data.

In the random fraction configuration, the exact proportion of anomalies in the test set is not known beforehand due to the random selection process. This fraction is determined after the data split based on the actual number of anomalies and normal instances present in the test set.

These configurations allow for an evaluation of the adaptability of anomaly detection algorithms and the associated evaluation metrics under different degrees of outlier prevalence, ranging from controlled to unpredictable scenarios.

Furthermore, we employed Spearman's rank correlation coefficient as a statistical measure to assess the dependency between the evaluation metrics~\cite{Schober2020Correlation}. This non-parametric measure assesses how well the relationship between two variables can be described using a monotonic function, providing insights into the consistency of the evaluation metrics across different contamination levels and outlier fractions~\cite{spearman04}. In the correlation analysis in our study, Spearman’s rank correlation coefficient was selected over Pearson's for its key advantages~\cite{spearman_pearson}. As a non-parametric measure, Spearman's method does not presuppose a linear relationship between variables. This is particularly pertinent to our research, where the relationships among evaluation metrics are potentially non-linear. Furthermore, Spearman's correlation exhibits robustness against outliers and non-normal distributions. This robustness is crucial as it ensures the reliability and accuracy of our metric analysis.

%For a comprehensive assessment of the evaluation metrics, we extend our experimental setup to include a simulated dataset. We generate Gaussian distributions for normal and anomalous instances, with varying degrees of mean separation. These simulations allow us to rigorously test the performance of evaluation metrics in a controlled manner.

For a comprehensive assessment of the evaluation metrics under controlled conditions, we extend our experimental setup to create a simulated environment. To this end, we generated  synthetic data representing normal and anomalous instances through Gaussian distributions. This approach enabled us to manipulate the degree of difficulty in distinguishing between normal and anomalous data by varying the mean separation between these distributions.

We defined two separate Gaussian distributions to model the normal and anomalous data points. Normal instances were generated from a Gaussian distribution with a mean of $0$ and a standard deviation of $1$, representing the baseline behavior or state. Anomalous instances were generated from a Gaussian distribution where the mean varied to represent different degrees of separation from the normal state, while maintaining the same standard deviation.

This setup was for testing how well each metric could identify anomalies as the distinction between normal and anomalous data varied. We chose a range of mean separations from $0$ to $5$, tested across $20$ evenly spaced intervals, to simulate scenarios ranging from highly overlapping (hardly distinguishable) to well-separated (easily distinguishable) distributions.

This experimental setup was designed to provide a comprehensive understanding of how different anomaly detection algorithms perform under varying conditions and how the chosen evaluation metrics respond to these changes. Our approach is designed to contribute to the broader discourse in the field, offering new insights into the applicability and reliability of these metrics in diverse anomaly detection scenarios.

\section{Methodology}
\subsection{Dataset Preparation}
\subsubsection{Selection and Characteristics}
Our study involved the use of $37$ datasets provided by the Anomaly Detection Benchmark (ADBench)~\cite{surveyzhao}. While the ADBench offers $47$ datasets, our choice was constrained to $37$ due to runtime considerations. These datasets represent diverse characteristics in terms of size, dimensionality, and types of anomalies. This range of diversity, despite not encompassing the entire ADBench collection, provides a comprehensive evaluation of anomaly detection methods.

% \subsubsection{Preprocessing Steps}
Normalization was applied to each dataset, aligning features to a mean of zero and scaling to unit variance. Normalization served to maintain consistency in preprocessing across datasets.

\subsection{Algorithm Implementation}
For the implementation of the KNN, LOF, OCSVM, and IForest algorithms in our experiments, we utilized the default parameters and implementation as provided by the pyod library~\cite{pyod}. This decision mirrors the typical application of these algorithms in practical scenarios. Our study is primarily focused on examining the general behavior and characteristics of these algorithms in various testing conditions, rather than optimizing for their peak performance. By using default settings, we aim to enhance the reproducibility and comparability of our results, thereby aligning our methodology with the objective of assessing the algorithms in realistic conditions.

\subsection{Training and Testing Split}
Two experimental setups were used to split the data into a training and a test set.
\subsubsection{Fixed Outlier Fraction}
In this setup, all anomalies were placed into the test set, establishing a $50\%$ outlier fraction alongside an equal number of normal observations. The rest of the data was allocated to train the models. This approach allows for a controlled evaluation environment with a known and fixed proportion of outliers.
\subsubsection{Random Split}
For the random split, $30\%$ of each dataset was randomly selected to form the test set, resulting in a variable outlier fraction for each dataset. This setup does not specifically aim for a $30\%$ anomaly proportion in the test set, but rather ensures a more realistic, almost random distribution of anomalies in the test set. This variability in the outlier fraction across datasets provides a more dynamic testing environment and mirrors real-world conditions more closely.

\subsection{Contamination Threshold}
The contamination threshold in our study is an essential parameter, representing the expected proportion of outliers in the training data. This threshold significantly impacts the sensitivity and specificity of anomaly detection algorithms and the resulting F1 score.

For the purposes of our study, we have set the contamination threshold at three distinct levels: $0.01$, $0.05$, and $0.1$. These levels were chosen to represent a range of scenarios. The rationale behind selecting these specific levels is to evaluate the adaptability and responsiveness of the anomaly detection algorithms and particularly to assess the impact on the F1 score under varying degrees of outlier presence.

These varying levels of contamination allow us to comprehensively assess the performance of the F1 score in different outlier conditions. The outcomes of this assessment are expected to offer valuable insights into the practical application of this metric in real-world anomaly detection tasks.

%The contamination threshold in our study is an essential parameter, representing the expected proportion of outliers in the training data. We set this threshold at three distinct levels: 0.01, 0.05, and 0.1, to represent a range of scenarios and their potential impact on anomaly detection algorithms.

%The primary focus of employing these varying thresholds is to examine how they influence the F1 score. This approach is intended to gauge the adaptability and responsiveness of the F1 score to different degrees of presumed outlier presence. By assessing the F1 score across these varied contamination levels, we aim to derive insights into its reliability and practical utility in real-world anomaly detection tasks.

\subsection{Evaluation Metrics}
The values used to calculate precision, recall, true positive rate (TPR) and false positive rate (FPR) come from confusion matrix. Table~\ref{tab:confusion_matrix} shows how the confusion matrix is structured~\cite{Ting2010}. %cite
\begin{table}[htbp]
\caption{Confusion matrix}
    \centering
    \begin{tabular}{|c|c|c|}
    \hline
         &  Predicted Positive & Predicted Negative\\
    \hline
     Actual Positive & True Positive (TP) & False Negative (FN) \\
     \hline
     Actual Negative & False Positive (FP) & True Negative (TN) \\
     \hline
    \end{tabular}
    \label{tab:confusion_matrix}
\end{table}

Precision measures the accuracy of the positive predictions~\cite{precision_recall}. It is defined as the ratio of true positives to the total number of instances predicted as positive: %cite
    \begin{equation}
        \text{Precision} = \frac{\text{TP}}{\text{TP}+\text{FP}}
    \end{equation}

Recall, also known as true positive rate, measures the model's ability to identify all true positives within a dataset~\cite{precision_recall}. It is defined as the ratio of true positives to the actual total number of positives~\cite{ROC}: %cite
    \begin{equation}
        \text{Recall} = \text{TPR} = \frac{\text{TP}}{\text{TP}+\text{FN}}
    \end{equation}

False positive rate measures the proportion of negative instances that are incorrectly classified as positive~\cite{ROC}. It is defined as the ratio of false positives to the total number of actual negative instances: %cite
    \begin{equation}
        \text{FPR} = \frac{\text{FP}}{\text{FP}+\text{TN}}
    \end{equation}

\subsubsection{F1 score}
The F1 score is the harmonic mean of precision and recall, offering a balanced measure of model accuracy, particularly where both false positives and false negatives are consequential. It is defined as:
    \begin{equation}
        F1 = 2 \times \frac{\text{precision} \times \text{recall}}{\text{precision} + \text{recall}}
    \end{equation}

The F1 score depends on a specific threshold setting to determine the binary classification of cases as normal or anomaly. The choice of this threshold influences both the precision and recall, thereby affecting the F1 score. This metric indicates the overall robustness of anomaly detection models in balanced class distributions~\cite{sasaki2007truth}.% It is of a family of metrics, the $F_\beta$ score%no like who cares7
\subsubsection{ROC AUC (Receiver Operating Characteristic Area Under Curve)}
ROC AUC assesses the likelihood of a random abnormal sample being more anomalous than a normal one. It is visualized as a curve plotting the true positive rate (TPR) against the false positive rate (FPR), and can be mathematically expressed as an integral across all thresholds. The ROC AUC value ranges from $0$ to $1$, where a higher value indicates better model discrimination~\cite{ROC}.
\subsubsection{AUCPR (Area Under the Precision-Recall Curve)}
AUCPR, designed for imbalanced datasets, emphasizes the minority class by plotting precision against recall across various thresholds. It evaluates the model's performance in scenarios where false negatives carry more weight than false positives, making it particularly useful in fields where the consequences of missing true positive cases are critical~\cite{10.1145/1143844.1143874}. Similar to ROC AUC, AUCPR can be mathematically expressed as an integral across the entire precision-recall curve~\cite{AUCPR_explain}. This integral representation provides a comprehensive assessment of the model's performance over various threshold levels, reflecting its effectiveness in distinguishing between classes in imbalanced situations.

\subsection{Correlation Analysis}
An integral component of our analysis was the implementation of a correlation analysis, specifically focusing on the relationships between evaluation metrics. This analysis aimed to quantitatively evaluate the consistency or divergence in the behaviors of these metrics across various scenarios.

To conduct this correlation analysis, we aggregated metric values from all datasets. We then applied Spearman’s rank correlation coefficient to assess the relationships between each pair of metrics~\cite{Schober2018Correlation}. 

\subsection{Simulation of Datasets}
We complement the analysis with simulated datasets to evaluate the behavior of anomaly detection metrics under controlled conditions. We generated synthetic datasets with predefined parameters to model the distinction between normal and anomalous data points. This approach enables us to evaluate the chosen metrics' sensitivity across a spectrum of anomaly detection scenarios, from clear to ambiguous separations between normal and anomalous instances.

A percentile threshold was determined for classifying scores into normal or anomalous, where the threshold corresponded to the 95th percentile of the combined score distribution (equivalent to a contamination level of $5\%$). The evaluation results were visualized using a line plot that compared the performance of the three metrics across the mean separation spectrum, providing a clear depiction of their responses to changes in the underlying data distribution.

\subsection{Software and Tools}
For outlier detection tasks, we utilized the Python Outlier Detection (PyOD) library~\cite{pyod}.

\section{Results and Discussion}
The experimental results demonstrate distinct correlation patterns among the F1 score, ROC AUC, and AUCPR across different contamination levels and outlier fractions. Initial observations from our correlation analysis indicate that when the outlier fraction in the test set is stable, the F1 score and AUCPR metrics demonstrate a significant and consistent correlation, with Spearman correlation coefficients ranging from $0.86$ to $0.90$. This high correlation persists across all contamination levels tested, suggesting a robust relationship between these two metrics under stable outlier conditions.

\begin{figure}[htbp]
\centerline{\includegraphics[width=\columnwidth]{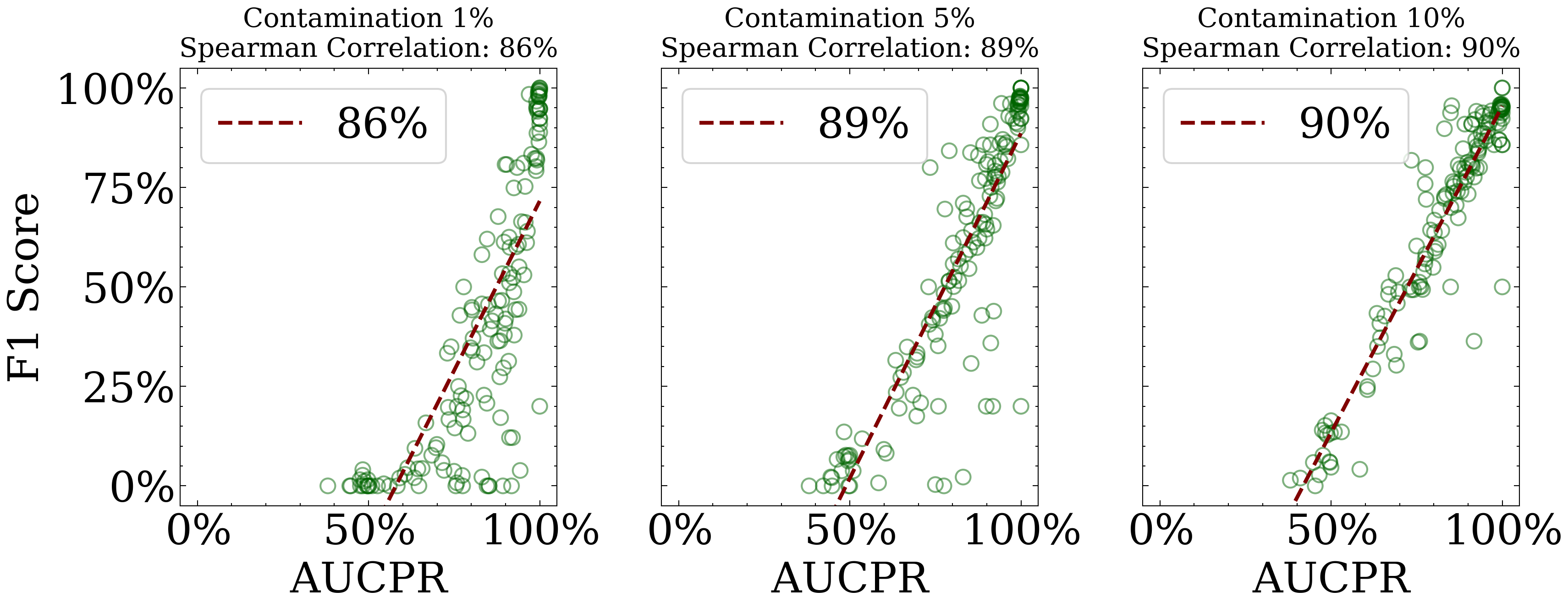}}
\caption{Correlation of F1 Score and AUCPR at fixed $50\%$ Outlier Fraction: This figure displays the strong Spearman correlation between the F1 score and AUCPR across contamination levels of $1\%$, $5\%$, and $10\%$. The correlation demonstrates the F1 score's increasing stability and decreasing variability with higher contamination levels, affirming a robust relationship between these metrics when the outlier fraction is stable.}
\label{fig_r1_F1_PR}
\end{figure}

Upon further examination, however, the F1 score exhibits considerable variability across different datasets as shown in Fig.~\ref{fig_r1_F1_PR}, which can be partly attributed to the single-threshold approach employed in its calculation. In scenarios with low contamination, the F1 score is more variable as only a few samples define the threshold, which leads to greater randomness. As contamination increases, this variability lessens, with more samples contributing to the threshold and leading to a more stable F1 score. This trend indicates that the F1 score becomes less random and more stable with higher contamination, explaining the observed increasing correlation between the F1 score and AUCPR.

In contrast, AUCPR, which is calculated directly from the model's scores without the need for a predetermined threshold, remains unaffected by changes in contamination levels. This underscores the drawback of the sensitivity of the F1 score to contamination levels, highlighting the need for careful interpretation of this metric, especially in environments with diverse or uncertain contamination rates. The F1 score's response to contamination levels, distinct from a specific threshold-independent metrics like AUCPR, necessitates a nuanced approach in its application for anomaly detection tasks.

\begin{figure}[htbp]
\centerline{\includegraphics[width=\columnwidth]{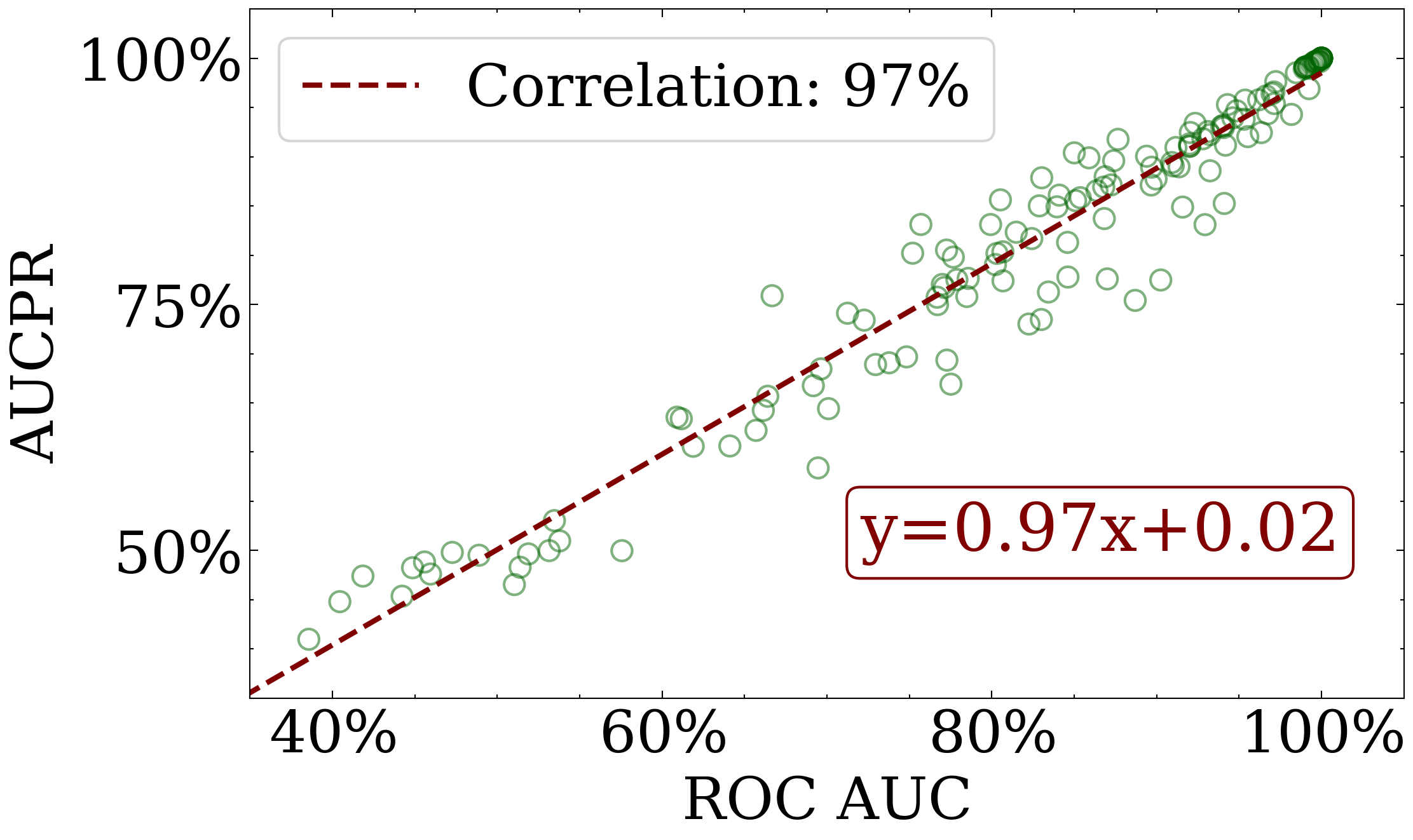}}
\caption{Alignment of AUCPR and ROC AUC at a stable outlier fraction of $50\%$: The figure illustrates the near-perfect alignment between AUCPR and ROC AUC with a $97\%$ Spearman correlation coefficient. This high correlation indicates that AUCPR is virtually equivalent to ROC AUC, with minimal deviation, under conditions of constant outlier fraction.}
\label{fig_r1_ROC_PR}
\end{figure}

%AUCPR correlates ROCAUC we already talked about. But it is actually more. As long as frac=0.5->AUCPR is basically = ROCAUC. (MAE is less than 1% difference)
Concurrently, when the outlier fraction is stable in the test set, as depicted in Fig.~\ref{fig_r1_ROC_PR}
, ROC AUC and AUCPR demonstrate a remarkable alignment in their assessment of model performance, with the correlation between these metrics reaching as high as $97\%$.  Specifically, as long as the outlier fraction is stable, the AUCPR metric is statistically almost equivalent to ROC AUC, with the Mean Absolute Error (MAE) between them being less than $1\%$.

This alignment suggests that in controlled environments with a known and constant proportion of outliers, the choice between these two metrics may not be as critical as previously thought. The similar results yielded by both ROC AUC and AUCPR under these conditions imply that selecting either metric could be equally effective for evaluating anomaly detection models.

\begin{figure}[htbp]
\centerline{\includegraphics[width=\columnwidth]{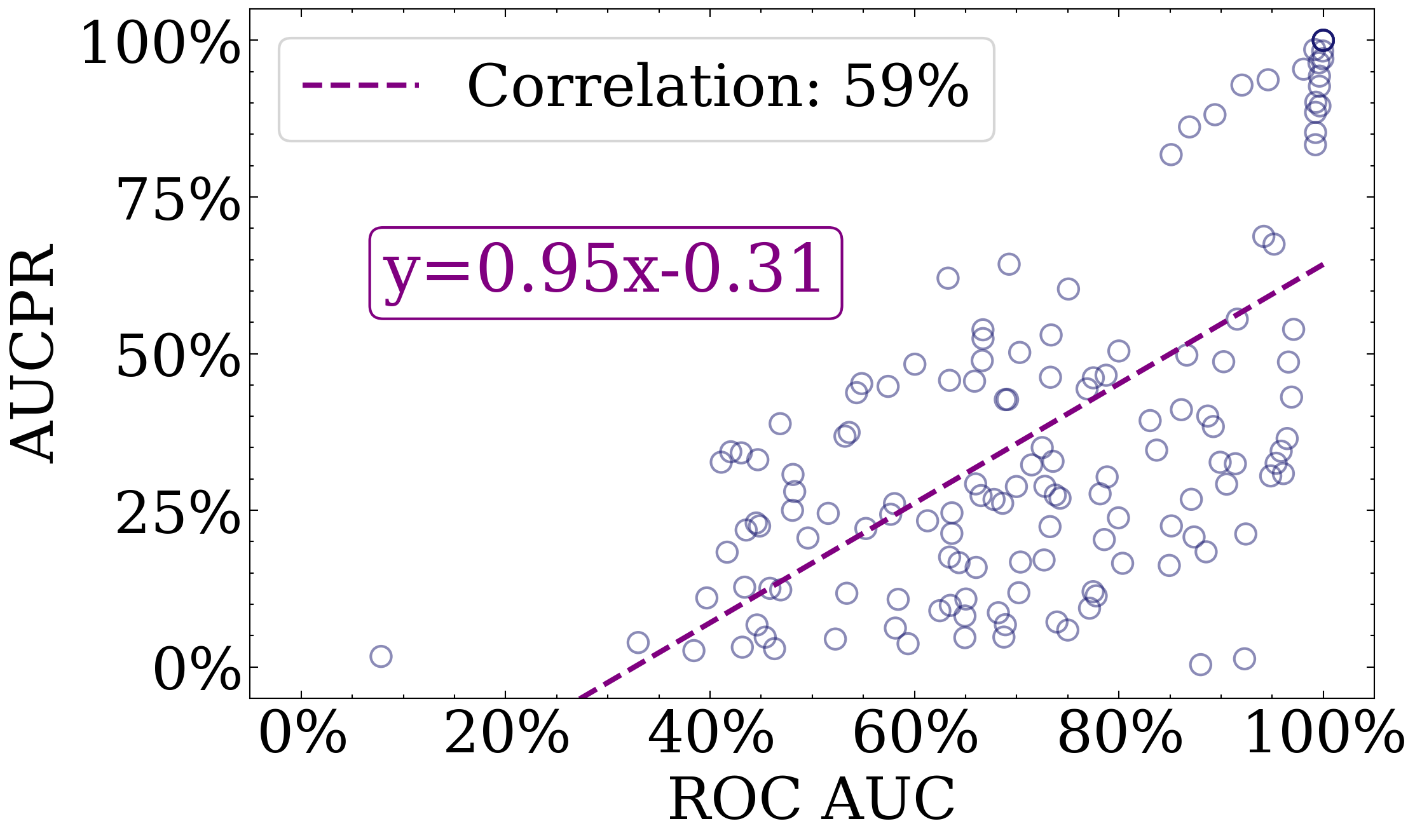}}
\caption{Low Correlation of AUCPR and ROC AUC in Variable Outlier Fractions: This figure illustrates the significant reduction in the correlation between AUCPR and ROC AUC when outlier fractions vary. It underscores the sensitivity of the AUCPR metric to changes in outlier distribution, emphasizing the challenges in performance evaluation for anomaly detection in environments with non-stable outlier conditions.}
\label{fig_r2_ROC_PR}
\end{figure}

Moving to variable outlier fractions, the correlation patterns shift significantly, as shown in Fig.~\ref{fig_r2_ROC_PR}. Examining the AUCPR metric correlation revealed a notable decrease when transitioning from stable to variable outlier fractions in the test set. Under conditions with stable outlier fractions, the AUCPR metric aligns closely with both the F1 score and ROC AUC, indicating a high level of agreement as shown in Fig.~\ref{fig_r1_F1_PR} and Fig.~\ref{fig_r1_ROC_PR}. However, this correlation significantly weakens under variable outlier conditions, a trend clearly depicted in Fig.~\ref{fig_r2_ROC_PR}. This observation underscores the AUCPR metric's sensitivity to fluctuations in outlier prevalence. Such variability in correlation complicates the performance evaluation process in anomaly detection, particularly in cases where the distribution of anomalies is irregular or unpredictable. These findings highlight the importance of considering the dynamic nature of outlier prevalence when interpreting the AUCPR metric in different anomaly detection scenarios.

\begin{figure}[htbp]
\centerline{\includegraphics[width=\columnwidth]{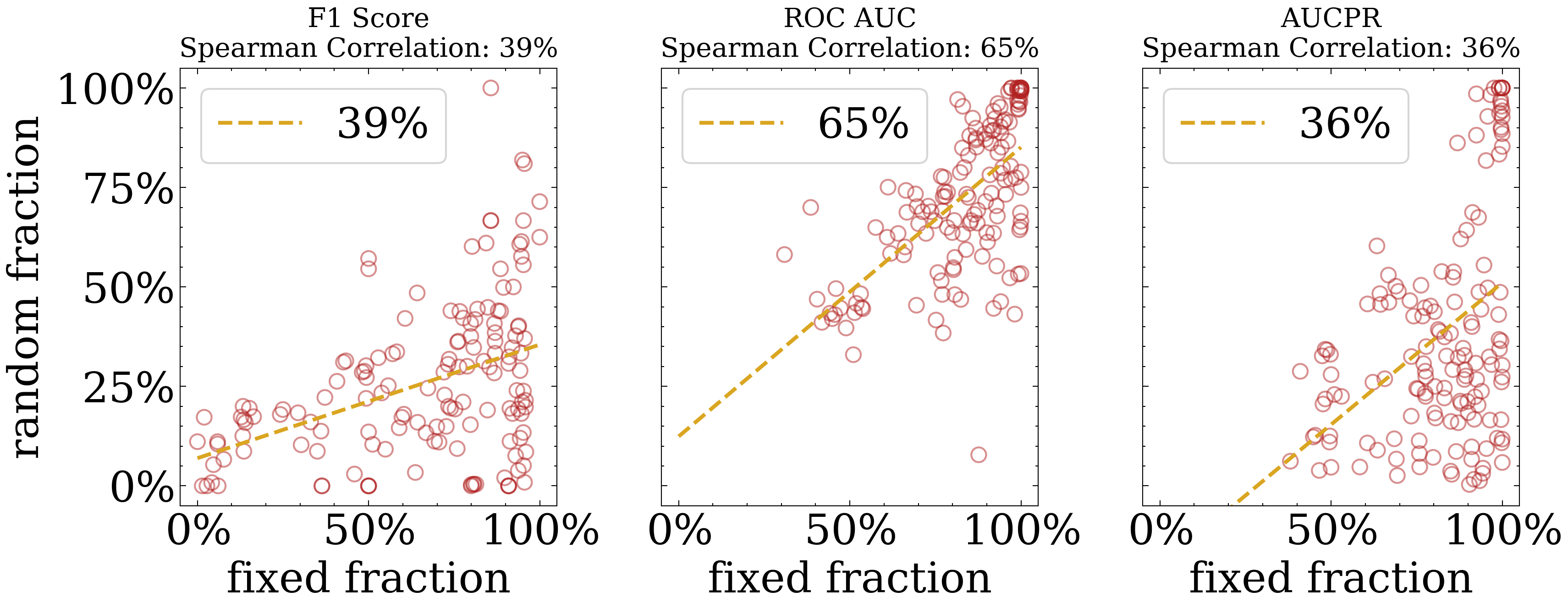}}
\caption{Comparative Analysis of F1 Score, ROC AUC, and AUCPR under Random and Fixed Fraction Conditions: This figure illustrates the variability and correlation of the F1 Score, ROC AUC, and AUCPR under both random and fixed outlier fractions. Notably, ROC AUC exhibits higher consistency and robustness across varying conditions, maintaining stronger correlations compared to F1 Score and AUCPR, which show significant variability, particularly under random fractions.}
\label{fig_vgl}
\end{figure}

In contrast to the F1 score and AUCPR, the ROC AUC metric's performance remains consistent and unaffected by the changes in contamination threshold and the variability of outlier fractions as shown in Fig.~\ref{fig_vgl}. The independence of the ROC AUC from these factors affirms its robustness, making it a reliable metric for evaluating anomaly detection models across a variety of experimental conditions.

In light of the ongoing debate about the utility of different evaluation metrics in anomaly detection, our results offer some clarifying insights, particularly concerning the choice between ROC AUC and AUCPR in specific experimental setups. As noted in the experimental design, the selection of the most appropriate metric often varies depending on individual preferences and the specific context of the study. However, our findings present an intriguing consistency between ROC AUC and AUCPR in scenarios with a stable outlier fraction.

\subsection{Simulated data}

%simulated data result
\begin{figure}[htbp]
\centerline{\includegraphics[width=\columnwidth]{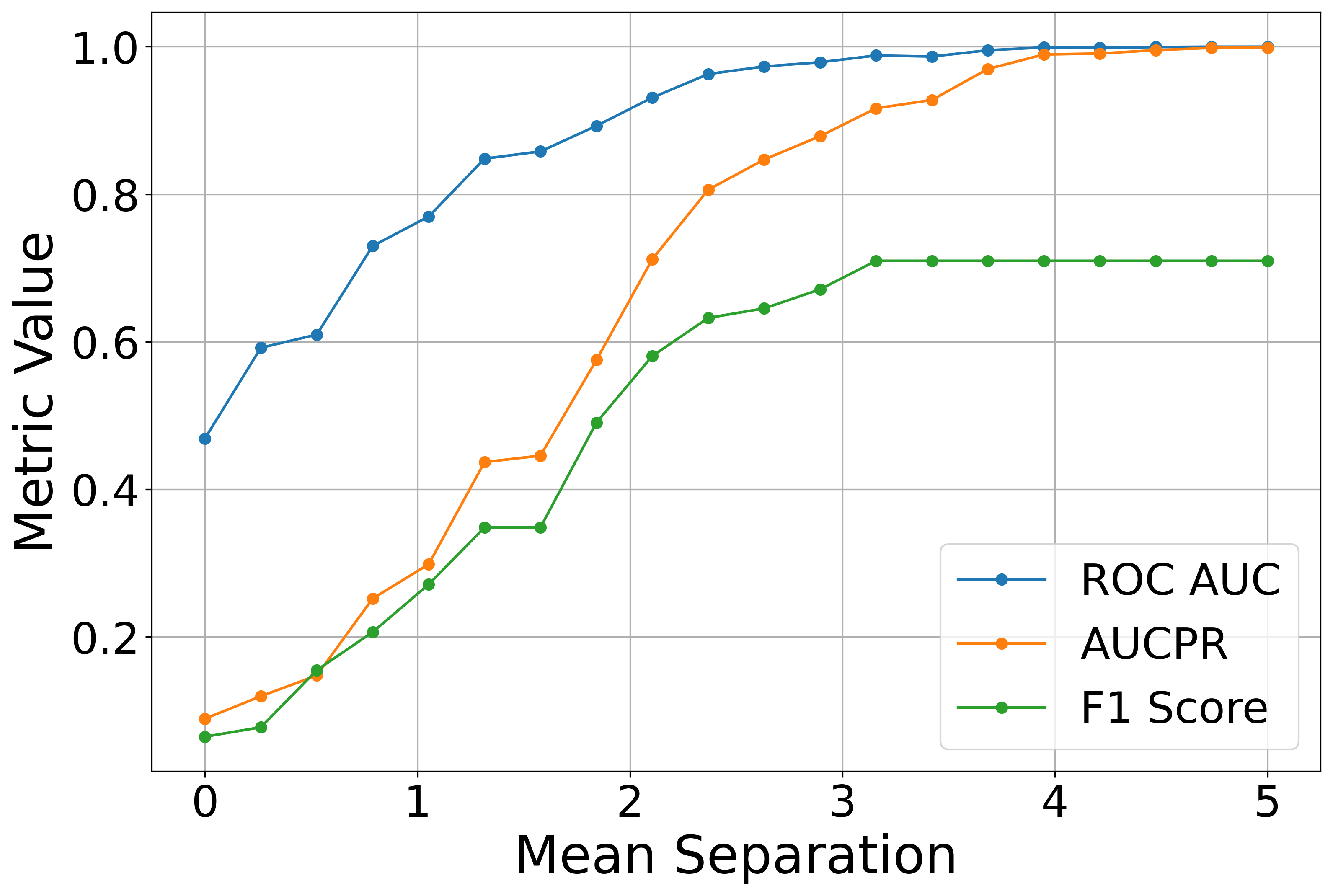}}
\caption{Comparative Performance of Evaluation Metrics with Increasing Mean Separation: Depicted here is how the ROC AUC and AUCPR metrics tend to converge as the mean separation between normal and anomalous distributions increases.}
\label{fig_gaussian_separation}
\end{figure}

In addition to our analysis on real-world datasets, we extended our investigation to include experiments with simulated data. The simulated data allowed us to control the mean separation between Gaussian distributions representing normal and anomalous data points, providing a clear picture of how the evaluation metrics respond under varying levels of detection difficulty.

The results from the simulated dataset are consistent with our findings from real-world datasets, as illustrated in Fig.~\ref{fig_gaussian_separation}. As the mean separation increased, illustrating a more distinct division between normal and anomalous instances, we observed a corresponding enhancement in the performance of the evaluation metrics.

Notably, as the mean separation increases, ROC AUC and AUCPR exhibit highly similar trends, with a Spearman's correlation of $100\%$. This observation, which is also reflected in our real-world dataset analysis, suggests that these metrics tend to converge under specific conditions, demonstrating an almost perfect alignment in their evaluation of model performance.

In summary, the analysis reveals that the F1 score and AUCPR are both affected by the outlier fraction, while only the F1 score is influenced by contamination levels. ROC AUC, on the other hand, demonstrates independence from these factors, maintaining consistency across varying conditions. %

\section{Conclusion}
This study has systematically explored the behavior of various evaluation metrics, particularly the F1 score, ROC AUC, and AUCPR, in the context of anomaly detection. Our results have illuminated nuanced differences in these metrics' performance under varying conditions, with a specific focus on outlier fractions and contamination levels. A critical finding of our research is the strong alignment between ROC AUC and AUCPR in scenarios with a fixed outlier fraction, indicating that these metrics can be used interchangeably in controlled environments. Based on these insights, we recommend the use of ROC AUC or AUCPR in situations where outlier fractions are stable and predictable.

Moreover, our investigation has revealed the F1 score's pronounced sensitivity to changes in contamination levels. This sensitivity necessitates a more considered approach to metric selection tailored to the dataset's unique characteristics. For scenarios with variable contamination levels or when precise threshold setting is challenging, we advise employing evaluation metrics that are independent of a single threshold, such as ROC AUC or AUCPR, to ensure a more reliable and robust assessment of model performance.

\subsection{Limitations}
Our study, while comprehensive, is not without limitations. The primary constraint lies in the scope of datasets used. While we employed a diverse range of datasets, they may not encompass all possible outlier score distributions encountered in real-world applications. However, to limit the impact of this constraint, we complemented our research with studies on simulated data, which allowed us to explore a wider range of conditions.

Furthermore, we acknowledge that our use of default parameters in the algorithms, though beneficial for reproducibility and ensuring a consistent baseline for comparison, may not represent the optimal performance achievable in each case. However, the extent to which the use of optimally tuned parameters would significantly impact our findings remains questionable.

% \subsection{Future Work}
% Another area for future exploration is the real-world applicability of these findings. While our results indicate a certain level of consistency between ROC AUC and AUCPR under fixed conditions, it remains to be seen how this translates to more dynamic, real-world datasets where outlier fractions are not as predictable.

Moreover, considering the rapid advancement in machine learning algorithms and techniques, ongoing research is needed to continuously evaluate and possibly refine these evaluation metrics. This includes assessing existing metrics and developing new ones that may offer improved or more nuanced insights into anomaly detection.

\section*{Acknowledgements}
This work was supported by the Lamarr-Institute for ML and AI, the Research Center Trustworthy Data Science and Security, the Federal Ministry of Education and Research of Germany and the German federal state of NRW. The Linux HPC cluster at TU Dortmund University, a project of the German Research Foundation, provided the computing power.

\bibliographystyle{IEEEtran}
\bibliography{references,new}

\end{document}